\documentclass[runningheads]{llncs}
\usepackage{booktabs}
\usepackage{courier}
\usepackage{bm}
\usepackage{graphicx}
\usepackage{makecell}
\usepackage{amsmath,amssymb} 
\usepackage{color}
\usepackage[pagebackref=true,breaklinks=true,letterpaper=true,colorlinks,bookmarks=false]{hyperref}

\begin{document}
\title{CTAP: Complementary Temporal Action Proposal Generation} 

\titlerunning{CTAP: Complementary Temporal Action Proposal Generation}
%
\author{Jiyang Gao$^\star$,
Kan Chen\thanks{indicates equal contribution. Code is in {\url{http://www.github.com/jiyanggao/CTAP}}.},
Ram Nevatia}
%
\authorrunning{Jiyang Gao$^\star$ \and Kan Chen$^\star$ \and Ram Nevatia}
%

\institute{University of Southern California\\
\email{\{jiyangga, kanchen, nevatia\}@usc.edu}}
\maketitle              
\begin{abstract}
Temporal action proposal generation is an important task, akin to object proposals, temporal action proposals are intended to capture ``clips'' or temporal intervals in videos that are likely to contain an action. Previous methods can be divided to two groups: sliding window ranking and actionness score grouping. Sliding windows uniformly cover all segments in videos, but the temporal boundaries are imprecise; grouping based method may have more precise boundaries but it may omit some proposals when the quality of actionness score is low. Based on the complementary characteristics of these two methods, we propose a novel Complementary Temporal Action Proposal (CTAP) generator. Specifically, we apply a Proposal-level Actionness Trustworthiness Estimator (PATE) on the sliding windows proposals to generate the probabilities indicating whether the actions can be correctly detected by actionness scores, the windows with high scores are collected. The collected sliding windows and actionness proposals are then processed by a temporal convolutional neural network for proposal ranking and boundary adjustment. CTAP outperforms state-of-the-art methods on average recall (AR) by a large margin on THUMOS-14 and ActivityNet 1.3 datasets. We further apply CTAP as a proposal generation method in an existing action detector, and show consistent significant improvements. 
\end{abstract}

\section{Introduction}\label{sec: intro}

We focus on the task of generating accurate temporal action proposals in videos; akin to object proposals for object detection~\cite{ren2015faster}, temporal action proposals are intended to capture ``clips'' or temporal intervals in videos that are likely to contain an action. There has been some previous work in this topic and it has been shown that, as expected and in analogy with object proposals, quality of temporal action proposals has a direct influence on the action detection performance~\cite{Gao_2017_ICCV,Shou_2016_CVPR}. High quality action proposals should reach high Average Recall (AR) with as few number of retrieved proposals as possible. 


\begin{figure}[t]
  \centering
    \includegraphics[width=0.95\textwidth]{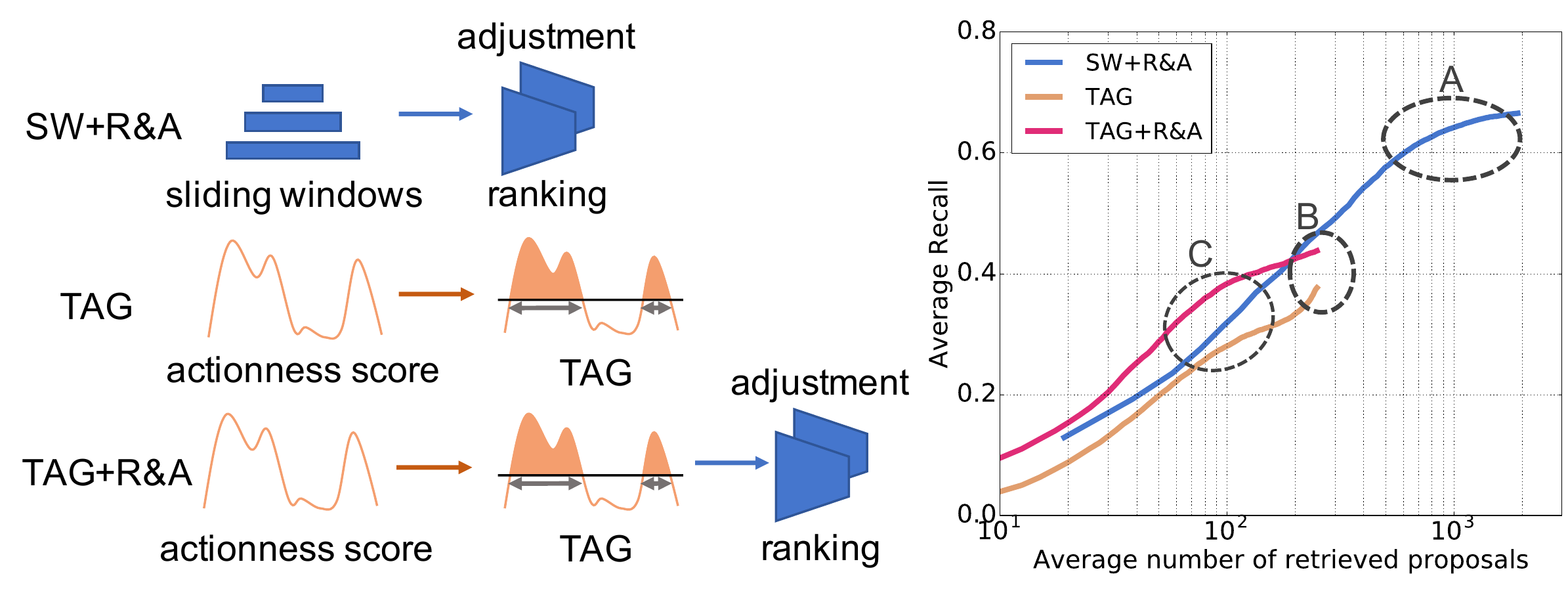}
    \caption{The architectures of three baseline methods are shown: (1) SW+R\&A: sliding windows are processed by a model for proposal ranking and boundary adjustment, \emph{e.g.} TURN\cite{Gao_2017_ICCV}, SCNN \cite{Shou_2016_CVPR}; (2) TAG: TAG \cite{Zhao_2017_ICCV} generate proposals based on unit-level actionness; (3) TAG+R\&A: actionness proposals are processed with proposal ranking and boundary adjustment.}
      \label{fig: baselines}
\end{figure}

The existing action proposal generation methods can be considered to belong to two main types. The first type is sliding-window based, which takes clips from sliding windows as input, and outputs scores for proposals. SCNN-prop~\cite{Shou_2016_CVPR} is a representative of this type; it applies a binary classifier to rank 
the sliding windows. TURN~\cite{Gao_2017_ICCV} adopts temporal regression in additional to binary classification to adjust the boundary of sliding windows. The architecture of this type is outlined as ``SW-R\&A'' in Fig.~\ref{fig: baselines}. 
Sliding windows uniformly cover all segments in the videos (thus cover every ground truth segment), however the drawback is that the temporal boundaries are imprecise, in spite of the use of boundary adjustment, and thus high AR is reached at large number of retrieved of proposals, as shown in circle A in Fig.~\ref{fig: baselines}.

The second type of action proposal generation methods can be summarized as actionness score based. It applies binary classification on a finer level, \emph{i.e.}, unit or snippet (a few contiguous frames) level, to generate actionness scores for each unit. 
A Temporal Action Grouping (TAG)~\cite{Zhao_2017_ICCV} technique, derived from the watershed algorithm \cite{roerdink2000watershed}, is designed to group continuous high-score regions as proposals. 
Each proposal's score is calculated as the average of its unit actionness scores. 
The structure is shown as ``TAG'' in Fig.~\ref{fig: baselines}. 
This type of method generates high precision boundaries, as long as the quality of actionness scores is high. 
However, the actionness scores have two common failure cases: having high scores at background segments, and having low scores at action segments. 
The former case leads to generation of wrong proposals, while the latter case may omit some correct proposals. These lead to the upper bound of AR performance limited at a low value (circle B in Fig.~\ref{fig: baselines}).

Based on the above analysis, ranking-sliding-window and grouping-actionness-score methods have two complementary properties: 
(1) The boundaries from actionness-based proposals are more precise as they are predicted on a finer level, and window-level ranking could be more discriminative as it takes more global contextual information; (2) actionness-based methods may omit some correct proposals when quality of actionness scores is low, sliding windows can uniformly cover all segments in the videos. 
Adopting the first complementary characteristic helps to resolve the first failure case of actionness proposals (\emph{i.e.}, generating wrong proposals). As shown in Fig.~\ref{fig: baselines}, a window-level classifier is applied after TAG to adjust boundaries and rank the proposals, which corresponds to model ``TAG+R\&A''.  Such combination has higher AR at low number of retrieved proposals compared to the sliding-window-based method (circle C in Fig.~\ref{fig: baselines}). However, it still fails to solve the second failure case, when actionness scores are low at true action segments, TAG is unable to generate these proposal candidates.
This results in the limited performance upper bound as shown in circle B, Fig.~\ref{fig: baselines}. 
To address this, we further explore the complementary characteristics, and propose to adaptively select sliding windows to fill the omitted ones in actionness proposals.

We propose a novel Complementary Temporal Action Proposal (CTAP) generator consisting of three modules. 
The first module is an initial proposal generator, which outputs actionness proposals and sliding-window proposals. 
The second module is a proposal complementary filter collects missing correct ones from sliding windows (addressing the second failure case of actionness score). Specifically, the complementary filter applies a binary classifier on the initial proposals to generate the probabilities indicating whether the proposals can be detected by actionness and TAG correctly, this classifier is called proposal-level actionness trustworthiness estimator.
The third module ranks the proposals and adjusts the temporal boundaries. Specifically, we design a temporal convolutional neural network, rather than simple temporal mean pooling used in TURN~\cite{Gao_2017_ICCV}, to preserve the temporal ordering information. 

We evaluated the proposed method on THUMOS-14 and ActivityNet v1.3; experiments show that our method outperforms state-of-the-art methods by a large margin for action proposal generation. 
We further apply the generated temporal proposals on the action detection task with a standard detector, and show significant performance improvements consistently. 

In summary, our contribution are three-fold:
(1) We proposed a novel Complementary Temporal Action Proposal (CTAP) generator which uses the complementary characteristics of actionness proposals and sliding windows to generate high quality proposals.
(2) We designed a new boundary adjustment and proposal ranking network with temporal convolution which can effectively save the ordering information on the proposal boundaries.
(3) We evaluated our method on two large scale datasets (THUMOS-14 and ActivityNet v1.3) and our model outperforms state-of-the-art methods by a large margin.

\section{Related Work}\label{sec: related work}
In this section, we introduce the related work, which includes temporal action proposal, temporal action detection and online action detection.

\textbf{Temporal Action Proposal.} Temporal action proposal generation has been shown to be an effective step in action detection, and could be useful for many high level video understanding tasks \cite{Shou_2016_CVPR,gao2017cascaded,Gao_2018_CVPR}. Two types of methods have been proposed, the first type of methods formulates it as a binary classification problem on sliding windows. Among them, Sparse-prop~\cite{Heilbron_2016_CVPR} uses STIPs~\cite{laptev2005space} and dictionary learning for proposal generation. SCNN-prop~\cite{Shou_2016_CVPR} is based on training C3D~\cite{tran2015learning} network for binary classification task. TURN~\cite{Gao_2017_ICCV} cuts the videos to units, and reuse unit-level features for proposals, which improves computational efficiency. TURN~\cite{Gao_2017_ICCV} also proposes to apply temporal regression to adjust the action boundaries which improves the AR performance. The performance of this type of methods is limited by the imprecise temporal boundaries of sliding windows. The second type of method is based on snippet level actionness score and apply Temporal Action Grouping (TAG)~\cite{Zhao_2017_ICCV} method on the score sequence to group continuous high-score region as proposal. However, TAG may omit the correct proposals when the quality of actionness scores is low. Besides, DAPs~\cite{escorcia2016daps} and SST~\cite{Buch_2017_CVPR} are online proposal generators, which could run over the video in a single pass, without the use of overlapping temporal sliding windows.

\textbf{Temporal Action Detection.} This task~\cite{Shou_2016_CVPR,Yeung_2016_CVPR,sun2015temporal,Yuan_2016_CVPR} focuses on predicting the action categories, and also the start/end times of the action instances in untrimmed videos. S-CNN~\cite{Shou_2016_CVPR} presents a two-stage action detection model, which first generates proposals and then classifies the proposals. Lin \emph{et al.} propose a Single Shot Action Detector (SSAD)~\cite{lin2017single}, which skips the proposal generation step and directly detects action instances in untrimmed video. Gao \emph{et al.}~\cite{gao2017cascaded} design a Cascaded Boundary Regression (CBR) network to refine the action boundaries iteratively. SSN~\cite{Zhao_2017_ICCV} presents a mechanism to model the temporal structures of activities, and thus the capability of discriminating between complete and incomplete proposals for precisely detecting actions. R-C3D~\cite{Xu_2017_ICCV} designs a 3D fully convolutional network, which generates candidate temporal regions and classifies selected regions into specific activities in a two-stage manner. Yuan \emph{et al.}~\cite{Yuan_2017_CVPR} propose to localize actions by searching for the structured maximal sum of frame-wise classification scores. Shou \emph{et al.} \cite{Shou_2017_CVPR} design a Convolutional-De-Convolutional (CDC) operation that makes dense predictions at a fine granularity in time to determine precise temporal boundaries. Dai \emph{et al.} \cite{Dai_2017_ICCV} propose a temporal context network, which adopts a similar architecture to Faster-RCNN \cite{ren2015faster}, for temporal action detection.
Beyond the fixed category action detection, TALL~\cite{Gao_2017_TALL} proposes to use natural language as the query to detect the target actions in videos.

Online action detection~\cite{de2016online,gao2017red,shou2018online} is different from temporal action detection that the whole video is not available at detection time, thus it needs the system to detect actions on the fly. Geest \textit{et al.}~\cite{de2016online} built a dataset for online action detection, which consists of 16 hours (27 episodes) of TV series with temporal annotation for 30 action categories. Gao \emph{et al.}~\cite{gao2017red} propose a Reinforced Encoder Decoder (RED) network for online action detection and action anticipation. 

\section{Complementary Temporal Action Proposal Generator}\label{sec: method}
In this section, we present the details of the Complementary Temporal Action Proposal (CTAP) generator. There are three stages in the pipeline of CTAP. The first stage is to generate initial proposals, which come from two sources, one is actionness score and TAG~\cite{Zhao_2017_ICCV}, the other is sliding windows. The second stage is complementary filtering. As we discussed before, TAG omits some correct proposals when the quality of actionness score is low (\emph{i.e.} low actionness score on action segments), but sliding windows uniformly cover all segments in videos. Thus, we design a complementary filter to collect high quality complementary proposals from sliding windows to fill the omitted actionness proposals. The third stage is boundary adjustment and proposal ranking, which is composed of a temporal convolutional neural network.


\begin{figure}[t]
  \centering
    \includegraphics[width=0.99\textwidth]{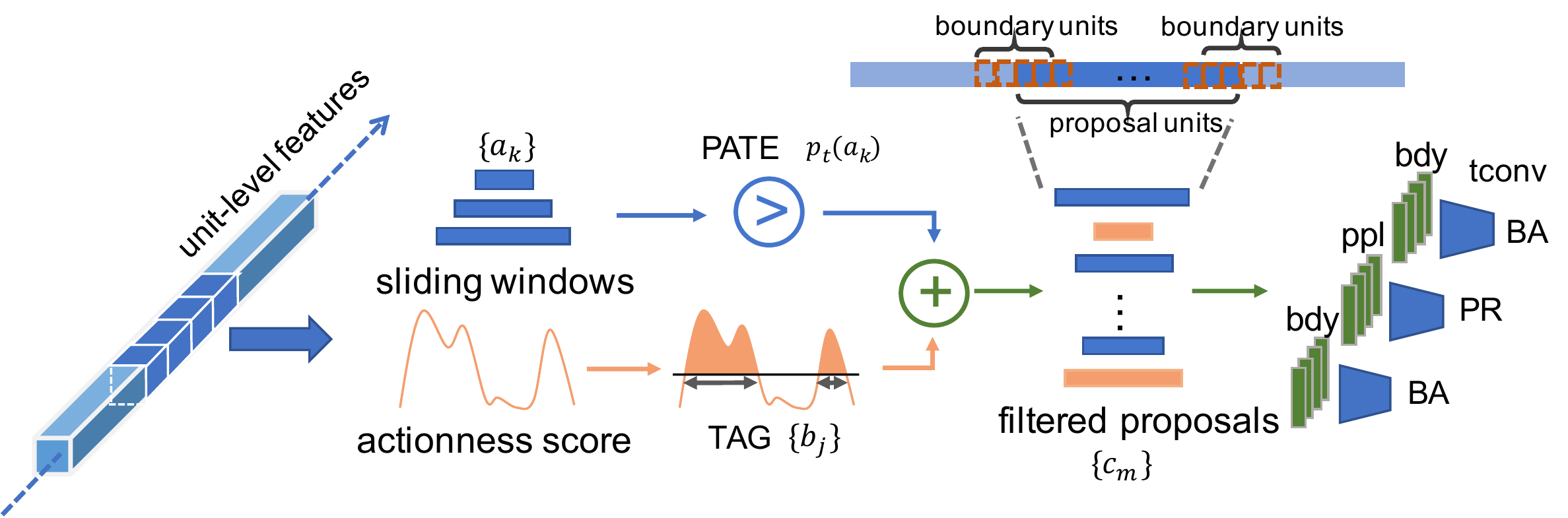}
    \caption{The architecture of Complementary Temporal Action Proposal (CTAP) generator. ``BA'' is short for boundary adjustment, ``PR'' is short for proposal ranking, ``ppl'' is short for proposal and ``bdy'' is short for boundary.}
      \label{fig: model}
\end{figure}

\subsection{Initial Proposal Generation}
In this part, we first introduce video pre-processing, then present the actionness score generation, temporal grouping process and sliding window sampling strategy. 

\textbf{Video pre-processing.} Following previous work~\cite{Gao_2017_ICCV}, a long untrimmed video is first cut into video units or snippets, each unit contains $n_u$ continuous frames. A video unit $u$ is processed by a visual encoder $E_v$ to extract the unit-level representation $\mathbf{x}_u=E_v(u)\in\mathbb{R}^{d_f}$. In our experiments, we use the two-stream CNN model~\cite{simonyan2014two,xiong2016cuhk} as the visual encoder, details are given in Sec~\ref{ssec: exp setup}. Consequently, a long video is converted to a sequence of unit-level features, which are used as basic processing units later.

\textbf{Actionness score}. Based on unit-level features, we train a binary classifier to generate actionness score for each unit. Specifically, we design a two-layer temporal convolutional neural network, which takes a $t_a$ continuous unit features as input, $\mathbf{x}\in\mathbb{R}^{t_a\times d_f}$, and outputs a probability for each unit indicating whether it is background or action, $\bm{p}_x\in\mathbb{R}^{t_a}$. 
\begin{equation}\label{equ: actioness score}
\bm{p}_x = \sigma(t_{conv}(\mathbf{x})),\text{\ \ \ } t_{conv}(\mathbf{x}) = \mathcal{F}(\varphi(\mathcal{F}(\mathbf{x}; \mathbf{W}_1));\mathbf{W}_2)
\end{equation}
where $\mathcal{F}(.;\mathbf{W})$ denotes a temporal convolution operator, $\mathbf{W}$ is the weight of its convolution kernel. 
In this network, $\mathbf{W}_1\in\mathbb{R}^{d_f\times d_m\times k \times k}, \mathbf{W}_2\in\mathbb{R}^{d_m\times1\times k \times k}$ ($k$ is the kernel size) are training parameters.
$\varphi(.)$ is an non-linear activation function, $\sigma(.)$ is a sigmoid function.

After generating the probability $\bm{p}_x$ for each continuous unit features $\mathbf{x}$, the loss is calculated as the cross-entropy for each input sample within the batch: 
\begin{equation}\label{equ: obj act score}
\mathcal{L}_{act} = -\frac{1}{N}\sum_{i=1}^N\left[\mathbf{y}_i^\top\log(\bm{p}_{x_i})+(1-\mathbf{y}_i)^\top\log(1-\bm{p}_{x_i})\right]
\end{equation}
where $\mathbf{y}_i\in\mathbb{R}^{t_a}$ is a binary sequence for each input $x_i$ indicating whether each unit in $x_i$ contains action (label 1) or not (label 0). $N$ is the batch size.

\textbf{Actionness proposal generation strategy}. We follow~\cite{Zhao_2017_ICCV} and implement a watershed algorithm \cite{roerdink2000watershed} to segment 1-D sequence signals. 
Given each unit's actionness score, raw proposals are generated whose units all have scores larger than a threshold $\tau$.
For some neighbor raw proposals, if the time during ration (\emph{i.e.}, maximum end time minus minimum start time among these raw proposals) is larger than a ratio $\eta$ of the whole video length, we group them as a proposal candidate. 
We iterate all possible combinations of $\tau$ and $\eta$ to generate proposal candidates and apply non-maximum suppression (NMS) to eliminate redundant proposals. 
The output actionness proposals are denoted as $\{b_j\}$.

\textbf{Sliding window sampling strategy.} Unlike actionness proposals which depend on actionness score distribution, sliding windows can uniformly cover all segments in the videos. The goal is to maximum the match with groundtruth segments (high recall), meanwhile maintaining the number of sliding windows as low as possible. In our experiments, different combinations of window size and overlap ratio are tested on validation set. The sliding windows are denoted as $\{a_k\}$. Detail setting is given in Sec~\ref{ssec: exp setup}.

\subsection{Proposal Complementary Filtering}

As discussed before, actionness proposals could be more precise but less stable, but sliding windows are more stable but less precise. The goal of second stage is to collect proposals, that could be omitted by TAG, from sliding windows. The core of this stage is a binary classifier, whose input is a sequence of unit features (\emph{i.e.} a proposal), and output is the probability that indicates whether this proposal can be correctly detected by the unit-level actionness scores and TAG. This classifier is called Proposal-level Actionness Trustworthiness Estimator (PATE).

\textbf{PATE training.} The training samples are collected as follows: Given a video, the groundtruth segments $\{g_i\}$ are matched with actionness proposals $\{b_j\}$. For a groundtruth segment $g_i$, if there exists an actionness proposal $b_j$ that has temporal Intersection over Union (tIoU) with $g_i$ larger than a threshold $\theta_c$, then we label $g_i$ as a positive sample ($y_i=1$); if no such $b_j$ exists, then $g_i$ is labelled as a negative sample ($y_i=0$). 
The unit level features inside $g_i$ are mean pooled to a single proposal-level feature $\mathbf{x}_{g_i}\in\mathbb{R}^{d_f}$. 
PATE outputs trustworthiness scores indicating the probabilities that whether the proposals can be correctly detected by actionness scores and TAG:
\begin{equation}\label{equ: pate score}
s_i = \sigma\left(\mathbf{W}_4(\varphi(\mathbf{W}_3\mathbf{x}_{g_i}+\mathbf{b}_3))+\mathbf{b}_4\right)
\end{equation}
where $\mathbf{W}_3\in\mathbb{R}^{d_f\times d_m}$, $\mathbf{W}_4\in\mathbb{R}^{d_m\times1}$, $\mathbf{b}_3\in\mathbb{R}^{d_m}, \mathbf{b}_4\in\mathbb{R}$ are training parameters. Other notations are similar to Eq.~\ref{equ: actioness score}. 
The network is trained by a standard cross-entropy loss over training samples from each batch ($N$ is the batch size).
\begin{equation}\label{equ: obj pate}
\mathcal{L}_{pate} = -\frac{1}{N}\sum_{i=1}^N\left[y_i\log(s_i)+(1-y_i)\log(1-s_i)\right]
\end{equation}

\textbf{Complementary filtering.} In test stage, we apply the trustworthiness estimator to every proposal from sliding windows $\{a_k\}$. For an input proposal, the trustworthiness score $p_t$ tells us that ``how well the actionness scores are trustworthy on the video content from this proposal''. For a sliding window $a_k$, if $p_t(a_k)$ is lower than a threshold $\theta_a$ (means TAG may fail on this segment), this sliding window is collected. The collected proposals from sliding windows and all actionness proposals are denoted as $\{c_m\}$, and are sent to the next stage, which ranks the proposals and adjusts the temporal boundaries. We call this process as complementary filtering and the name derives from somewhat similar processes used in estimation theory~\footnote{The original use of \href{https://ocw.mit.edu/courses/aeronautics-and-astronautics/16-333-aircraft-stability-and-control-fall-2004/lecture-notes/lecture_15.pdf}{complementary filtering} is to estimate a signal given two noisy measurements, where one of the noise is mostly high-frequency (maybe precise but not stable) and the other noise is mostly low-frequency (stable but not precise).}.

\subsection{Proposal Ranking and Boundary Adjustment}
The third stage of CTAP is to rank the proposals and adjust the temporal boundaries. TURN~\cite{Gao_2017_ICCV} does this also, however it uses mean-pooling to aggregate temporal features, which losses the temporal ordering information. Instead, we design a Temporal convolutional Adjustment and Ranking (TAR) network which use temporal conv layers to aggregate the unit-level features. 

\textbf{TAR Architecture.} Suppose that the start and end units (\emph{i.e.} temporal boundary) of an input proposal $c_m$ are $u_s, u_e$, we uniformly sample $n_{ctl}$ unit-level features inside the proposal, called proposal units $\mathbf{x}_c\in\mathbb{R}^{n_{ctl}\times d_f}$. We sample $n_{ctx}$ unit features at the start and end boundaries respectively, which are $[u_s-n_{ctx}/2, u_s+n_{ctx}/2]$ and $[u_e-n_{ctx}/2, u_e+n_{ctx}/2]$, called boundary units (denoted as $\mathbf{x}_s\in\mathbb{R}^{n_{ctx}\times d_f}, \mathbf{x}_e\in\mathbb{R}^{n_{ctx}\times d_f}$). 
Boundary units and proposal units are illustrated in Fig.~\ref{fig: model}. These three feature sequences (one sequence for proposal units and two sequences for boundary units) are input to three independent sub-networks. The proposal ranking sub-network outputs probability of action, the boundary adjustment sub-network outputs regression offsets. 
Each sub-network contains two temporal convolutional layers. which can be represented as:
\begin{equation}\label{equ: tar network}
o_s = t_{conv}(\mathbf{x}_s),\text{\ \ \ }p_c = \sigma(t_{conv}(\mathbf{x}_c)),\text{\ \ \ }o_e = t_{conv}(\mathbf{x}_e)
\end{equation}
where $o_s, o_e, p_c$ denote the offsets prediction for start and end boundaries and the action probability for each proposal respectively.
Other notations are the same in Eq.~\ref{equ: actioness score}.
Similar to TURN~\cite{Gao_2017_ICCV}, we use non-parameterized regression offsets. The final score for a proposal $a_k$ from sliding windows is multiplied by the PATE score ($p_t(a_k)\cdot p_c(a_k)$). The actionness proposals use $p_c(a_k)$ as the final score.

\textbf{TAR Training.} To collect training samples, we use dense sliding windows to match with groundtruth action segments. A sliding window is assigned to a groundtruth segments if: (1) it has the highest tIoU overlaps with a certain groundtruth segment among all other windows; or (2) it has tIoU larger than 0.5 with any one of the groundtruth segments. We use the standard Softmax cross-entropy loss to train proposal ranking sub-network and the L1 distance loss for boundary adjustment sub-network. Specifically, the regression loss can be expressed as,
\begin{equation}
\mathcal{L}_{reg}=\frac{1}{N_{pos}}\sum_{i=1}^{N_{pos}}l_i^*(|o_{s,i}-o^*_{s,i}| +|o_{e,i}-o^*_{e,i}|)
\end{equation}
where $o_{s,i}$ is the predicted start offset, $o_{e,i}$ is the predicted end offset, $o^*_{s,i}$ is the groundtruth start offset, $o^*_{e,i}$ is the groundtruth end offset. $l^*_i$ is the label, $1$ for positive samples and $0$ for background samples. $N_{pos}$ is the number of positive samples in a mini-batch, as the regression loss is calculated only for positive samples. Similar to Eq.~\ref{equ: obj pate}, a cross entropy objective is calculated to guide the learning of prediction score $p_c$ for each proposal.

\section{Experiments}\label{sec: exp}
We evaluate CTAP on THUMOS-14~\cite{THUMOS14} and ActivityNet v1.3~\cite{caba2015activitynet} datasets respectively.

\setlength{\parindent}{0pt}
\subsection{Datasets}

\textbf{THUMOS-14} contains 1010 and 1574 videos for validation and testing purposes from 20 sport classes. 
Among them, there are 200 and 212 videos are labeled with temporal information in validation and test set respectively. 
Following the settings of previous work~\cite{Gao_2017_ICCV,Shou_2016_CVPR}, we train our model on the validation set and conduct evaluation on the test set. 

\textbf{ActivityNet v1.3} consists of 19,994 videos collected from YouTube labeled in 200 classes. 
The whole dataset is divided into three disjoint splits: training, validation and test, with a ration of 50\%, 25\%, 25\%, respectively. 
Since the annotation of the test split is not publicly available for competition purpose, we compare and report performances of different models on the validation set. 

\subsection{Experiment Setup}\label{ssec: exp setup}
\textbf{Unit-level feature extraction.} We use the twostream model~\cite{xiong2016cuhk} as the visual encoder $E_v$ that is pre-trained on ActivityNet v1.3 training set. In each unit, the central frame is sampled to calculate the appearance CNN feature, it is the output of \texttt{Flatten\_673} layer in ResNet~\cite{He_2016_CVPR}. For the motion feature, we sample 6 consecutive frames at the center of a unit and calculate optical flows between them; these flows are then fed into the pretrained BN-Inception model~\cite{ioffe2015batch} and the output of \texttt{global pool} layer is extracted. The motion features and the appearance features are both 2048-dimensional, and are concatenated into 4096-dimensional vectors ($d_f=4096$), which are used as unit-level features. On THUMOS-14, we test our model with two settings of unit features Flow-16 and Twostream-6. Flow-16 only uses denseflow CNN features, and the unit size is set to 16, which is the same as~\cite{Gao_2017_ICCV}($n_u=16$), Twostream-6 use two-stream features and unit size is 6 ($n_u=6$). On ActivityNet v1.3, two-stream features are used and unit size is 16 (Twostream-16, $n_u=16$).

\textbf{Sliding window sampling strategy.} We follow TURN~\cite{Gao_2017_ICCV} and adopt proposals' length set of \{16, 32, 64, 128, 256, 512\} with tIOU of 0.75, which achieves the optimal results. On ActivityNet v1.3, we adopt proposals' length set of \{64, 128, 256, 512, 768, 1024, 1536, 2048, 2560, 3072, 3584, 4096, 6144\} with tIOU = 0.75, which achieves the reported best performance in the submission.

\textbf{Actionness score generation.} We set the kernel size for each temporal convolution as $3$ ($k=3$). The stride for temporal convolution is $1$. We choose rectified linear unit (ReLU) as the non-linear activation function $\varphi$. The first temporal convolution output dimension $d_m = 1024$. $t_a$ is set to be 4. Batch size is 128, learning rate is 0.005, and the model is trained for about 10 epochs. 

\textbf{TAG algorithm.} Following the setting of~\cite{Zhao_2017_ICCV}, we set the initial value of $\tau$ as $0.085$. To enumerate all possible combinations of $(\tau, \eta)$, we first iterate $\tau$ in the range of $[0.085, 1)$ with a step of $0.085$. In each iteration, we further iterate $\eta$ in the range of $[0.025, 1]$ with a step of 0.025. 
The threshold of NMS is set as $0.95$ to eliminate redundant proposals.

\textbf{PATE setting.} We set the first fully-connected layer's output dimension $d_m = 1024$. $\theta_a$ is set to be 0.1 on THUMOS-14 and ActivityNet v1.3. Batch size is 128 and learning rate is 0.005. PATE is trained for about 10 epochs.

\textbf{TAR setting.} On THUMOS-14, we uniformly sample $8$ unit features inside each proposal ($n_{ctl}=4$), and $4$ unit features as context ($n_{ctx} = 4$). 
On ActivityNet v1.3, we set $n_{ctl}=8$ and $n_{ctx} = 4$. $d_m$ is set to 1024.
TAR is optimized using Adam algorithm~\cite{kingma2014adam}. Batch size is 128 and learning rate is 0.005. TAR is trained for 10 epoches on THUMOS-14 and 4 epoches on ActivityNet v1.3.

\textbf{Evaluation Metrics.} For temporal action proposal generation task, Average Recall (AR) is usually used as evaluation metrics. Following previous work, we use IoU thresholds set from 0.5 to 1.0 with a step of 0.05 on THUMOS-14 and 0.5 to 0.95 with a step of 0.05 on ActivityNet v1.3. We draw the curve of AR with different Average Number(AN) of retrieved proposals to evaluate the relationship between recall and proposal number, which is called AR-AN curve. On ActivityNet v1.3, we also use area under the AR-AN curve (AUC) as metrics, where AN varies from 0 to 100. 
For the evaluation of temporal action detection, we follow the traditional mean Average Precision (mAP) metric
used in THUMOS-14. A prediction is regarded as positive only when it has correct category prediction and tIoU with ground truth higher than a threshold. We use the official toolkit of THUMOS-14.


\subsection{Evaluation on THUMOS-14}
In this part, we evaluate our method on THUMOS-14 dataset. First, we compare our proposal ranking and boundary adjustment module TAR with TURN~\cite{Gao_2017_ICCV}. Second, we evaluate the effectiveness of PATE and the proposal complementary filtering module. Third, we compare our full model with state-of-the-art methods, and finally we apply our proposals on action detection task to verify the its performance advantage.

\begin{table}
\centering
\caption{Performance comparison between TAR and TURN \cite{Gao_2017_ICCV} on THUMOS-14 test set. Same unit feature (flow-16) and test sliding windows are used on TAR and TURN for fair comparison. Average Recall (AR) at different numbers is reported.}
\label{tbl: TAR}
\begin{tabular}{l|cccc}
\toprule
 Method &\ AR@50\ &\ AR@100\ &\ AR@200\ &\\ \midrule
TURN\cite{Gao_2017_ICCV}\ \ &\ 21.75\ &\ 31.84\ &\ 42.96 \\ 
TAR &\ \textbf{22.99}\ &\ \textbf{32.21}\ &\ \textbf{ 45.08}\\ \bottomrule
\end{tabular}
\end{table}

\textbf{TAR vs TURN~\cite{Gao_2017_ICCV}.} As we presented before, TURN~\cite{Gao_2017_ICCV} uses temporal mean pooling to aggregate features, it losses temporal ordering information, which could be important for boundary adjustment. TAR uses temporal convolution to extract temporal information from unit features, and adopts independent sub-networks for proposal ranking and boundary adjustment. To fairly compare with TURN, we use flow-16 features, and the same test sliding window settings as TURN. As shown in Table~\ref{tbl: TAR}, we can see that, at AN=50, 100 and 200, TAR outperforms TURN at all these points, which shows the effectiveness of TAR.

\begin{table}[b]
\centering
\caption{Complementary filtering evaluation on THUMOS-14 test set, compared with ``Union'' and ``tIoU-selection''. Average Recall (AR)  at different numbers is reported.}
\label{tbl: fusion}
\begin{tabular}{l|cccc}
\toprule
Method &\ AR@50\ &\ AR@100\ &\ AR@200\ \\ \midrule
Union & \ 25.80\ &\ 34.70\ &\ 46.19\ \\ 
Union+NMS & \ 28.07\ &\ 39.71\ &\ 49.60\ \\ 
tIoU-selection &\ 30.35\ &\ 38.34\ &\ 42.41\ \\
PATE complementary filtering\ \ \ &\ \textbf{31.03}\ &\ \textbf{40.23}\ &\ \textbf{50.13}\ \\ \bottomrule
\end{tabular}
\end{table}

\textbf{Complementary filtering.} Besides using PATE in the proposal complementary filtering, we design three baseline methods to combine the sliding windows and actionness proposals. The first method is a simple ``union'', in which we simply put all actionness proposals and all sliding windows together, and send them into TAR module for ranking and adjustment. The second method is ``union''+NMS, in which we apply NMS to filter the duplicate proposals from the union set; the threshold of NMS is set to 0.7, which achieves the best performance among \{0.5,0.7,0.9\}. The third method is tIoU-based: all actionness proposals are selected; we calculate the tIoU between the sliding windows and actionness proposals, if there exists a sliding window whose highest tIoU with all actionness proposals is lower than 0.5, then it is selected.  We use flow-16 unit features and the same test sliding windows in ``TAR vs TURN'' experiments. 

The results are shown in Table~\ref{tbl: fusion}. We can see that, complementary filtering achieves the best AR on every AN (50, 100 and 200). The performance of ``Union'' suffers at low AN, but is higher than ``tIoU-selection'' at AN=200. We believe the reason is that simple union method adds too many low quality proposals from sliding windows. Union+NMS improves the performance, however due to the lack of priority of TAG and SW proposals, NMS may select an inaccurate SW proposal with a higher score instead of an accurate TAG proposal with a lower score. In contrast, PATE tries to preserve such priority and focuses on picking out the sliding window proposals that TAG may fail on. tIoU-selection also suffers, as it eliminates some high quality windows simply based on the tIoU threshold. Complementary filtering dynamically generates trustworthiness scores on different windows, which make the selection process more effective.

\begin{figure}[t]
  \centering
    \includegraphics[width=0.93\textwidth]{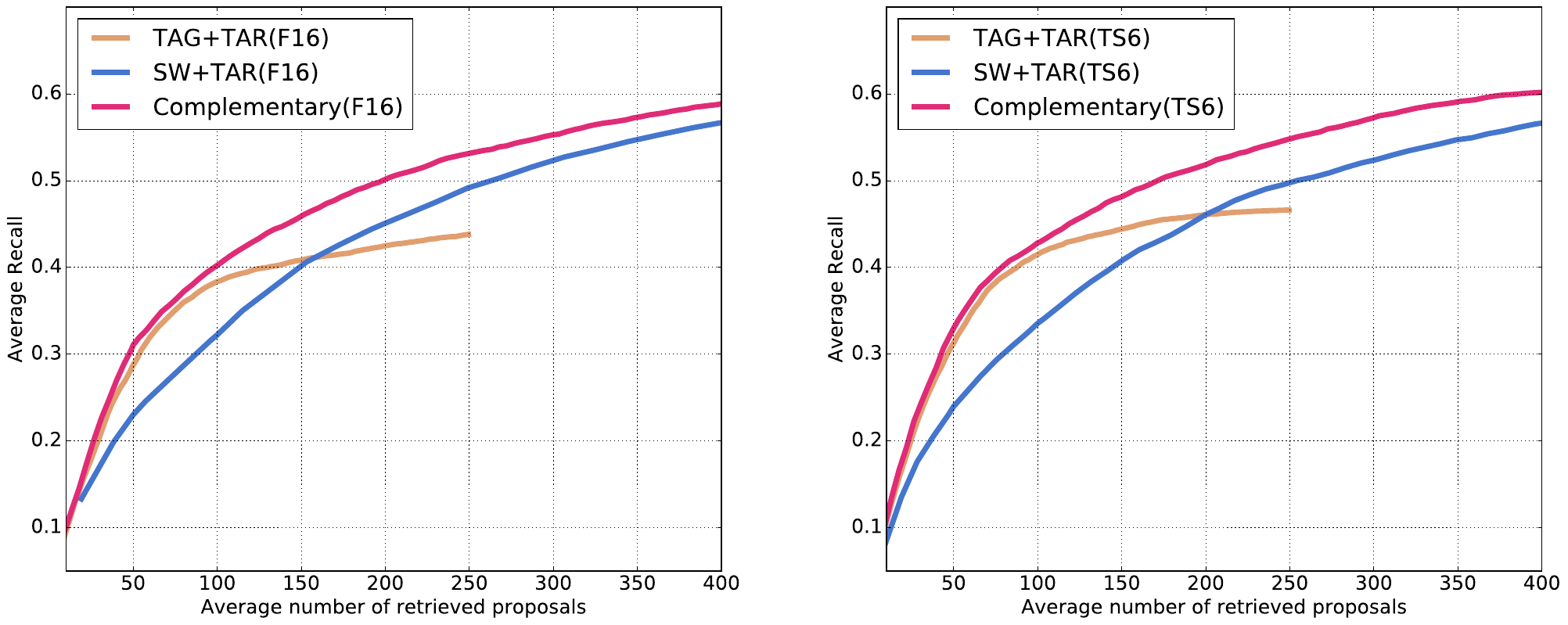}
    \caption{AR-AN curves of the complementary results with flow-16 feature (F16) and two-stream-6 feature (TS6). Complementary filtering proposals outperform sliding windows (SW+TAR) and actionness proposals (TAG+TAR) consistently.}
      \label{fig: fusioncomp}
\end{figure}

We also show the AR performance of two sources, actionness proposals and sliding windows, in Fig.~\ref{fig: fusioncomp}. Both flow-16 (F16) feature and twostream-6 (TS6) feature are illustrated. It can be seen that the performance of complementary proposals is higher than that of actionness proposals (TAG+TAR) and sliding windows (SW+TAR) at every AN consistently, which shows that our method can effectively select high quality complementary proposals from sliding windows to fill the omitted ones in actionness proposals.

\begin{figure}[t]
  \centering
    \includegraphics[width=0.93\textwidth]{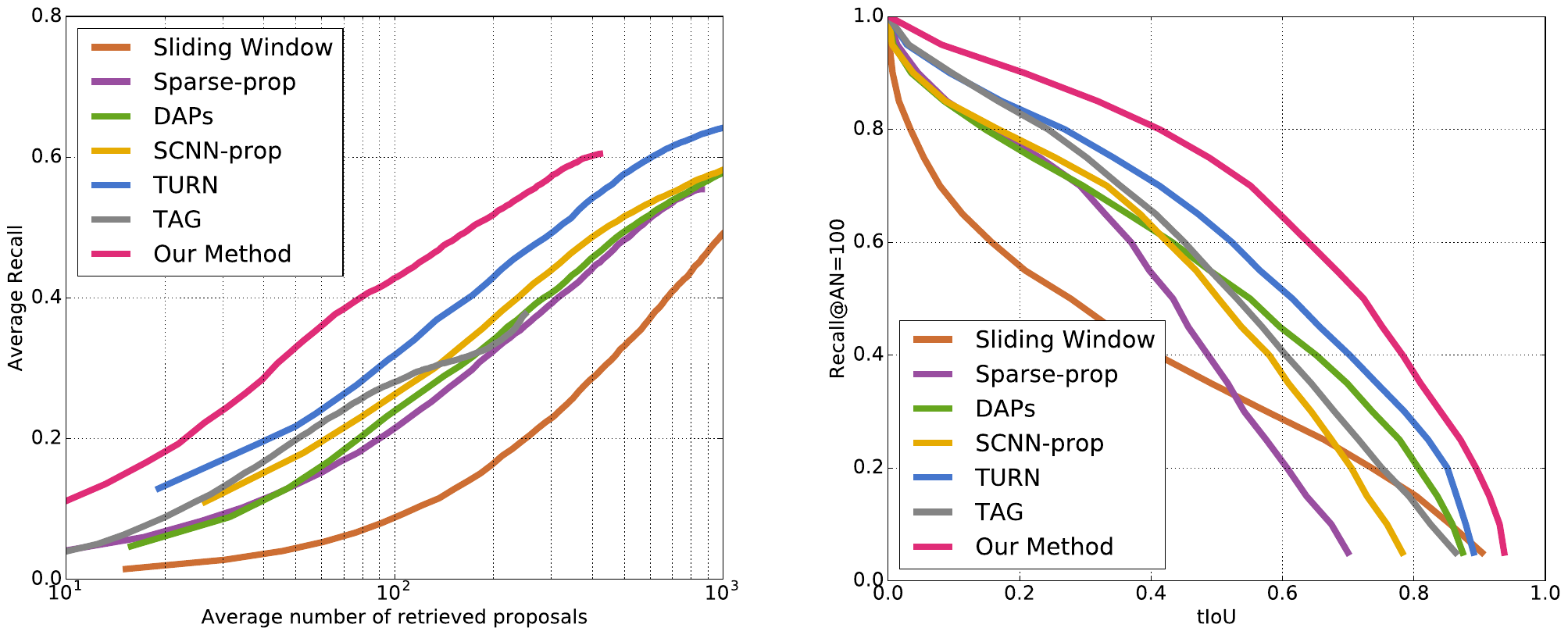}
    \caption{AN-AR curve and recall@AN=100 curve of CTAP and state-of-the-art methods on THUMOS-14 test set.}
      \label{fig: stoa}
\end{figure}

\textbf{Comparison with state-of-the-art methods.} We compare our full model with state-of-the-art methods on THUMOS-14 dataset by the Average recall on average number of proposals (AR-AN) curve and recall@100-tIoU curve, as shown in Fig.~\ref{fig: stoa}. It can be seen that our model outperforms the state-of-the-art model by a large margin on both curves. Specifically, for AR@100, the performance of CTAP is around 43\%, while the state-of-the-art method TURN \cite{Gao_2017_ICCV} only achieves about 32\%.

\textbf{CTAP for Temporal action detection.} To verify the quality of our proposals, we feed CTAP proposals into SCNN~\cite{Shou_2016_CVPR}, and compare with other proposal generation methods on the same action detector (SCNN). The results are shown in Table~\ref{tbl: detection}. We can see that our CTAP-TS6 achieves the best performance, and outperforms the state-of-the-art proposal method TURN~\cite{Gao_2017_ICCV} and TAG~\cite{Zhao_2017_ICCV} by over 4\%, which proves the effectiveness of the proposed method. 

\begin{table}[t]\scriptsize
\centering
\caption{Comparison of CTAP and other proposal generation methods with the same action detector (SCNN) on THUMOS-14 test set, mean Average Precision (mAP \% @tIoU=0.5) is reported. }
\label{tbl: detection}
\begin{tabular}{l|*5c|cc}
\toprule
 Method\ \ &\ Sparse~\cite{Heilbron_2016_CVPR}\ &\ DAPs~\cite{escorcia2016daps}\ &\ SCNN-prop\cite{Shou_2016_CVPR}\ &\ TURN~\cite{Gao_2017_ICCV}\ &\ TAG\cite{Zhao_2017_ICCV}\ \ &\ CTAP-F16\ &\ CTAP-TS6\ \\ \midrule
tIoU=0.5\ \ &\ 15.3\ &\ 16.3\ &\ 19.0\ &\ 25.6\ &\ 25.9\ \ &\ 27.9\ &\ \textbf{29.9}\ \\ \bottomrule
\end{tabular}
\end{table}

\begin{table}[b]
  \centering
  \caption{Evaluation of TURN~\cite{Gao_2017_ICCV}, TAR, MSAR~\cite{yao2017msr}, Prop-SSAD~\cite{lin2017temporal} and CTAP on ActivityNet v1.3 validation set. AR@100 and AUC of AR-AN curve are reported. (The AR@100 of MSRA~\cite{yao2017msr} is not available.)}\label{tbl: anet tar eval}
  \begin{tabular}{l|cc|cc|cc|c} \toprule
  Method\ & \makecell{SW-\\\ \ TURN~\cite{Gao_2017_ICCV}\ \ } & \makecell{TAG-\\\ \ TURN~\cite{Zhao_2017_ICCV}\ \ } & \makecell{SW-\\\ \ TAR\ \ } & \makecell{\ \ TAG-\ \ \ \\\ \ TAR\ \ \ } & \makecell{\ \ MSRA\ \ \\\cite{yao2017msr}} & \makecell{\ Prop-\\\ \ SSAD~\cite{lin2017temporal}\ \ } &\ CTAP\ \ \\ \midrule
  AR@100\ \ &\ 49.73\ &\ 63.46\ \ &\ 68.02\ &\ 64.01\ &\ -\ &\ 73.01\ &\ \textbf{73.17}\ \\
  AUC\ &\ 54.16\ &\ 53.92\ \ &\ 61.02\ &\ 64.62\ &\ 63.12\ &\ 64.40\ &\ \textbf{65.72}\ \\
  \bottomrule
  \end{tabular}
\end{table}

\subsection{Evaluation on ActivityNet v1.3}



\textbf{Evaluation of TAR.} To show the effectiveness of TAR, we report the AR@100 values and area under AR-AN curve for different models in Table~\ref{tbl: anet tar eval}. 
For sliding window proposals, we observe that TAR's prediction (SW-TAR) achieves 18.29\% and 6.86\% improvement in AR@100 and AUC compared to those of TURN~\cite{Gao_2017_ICCV} (SW-TURN). The results show that TAR is more effective in temporal boundary adjustment and proposal ranking. For actionness proposals, we observe that TAR achieves 10.70\% increase compared to TURN~\cite{Gao_2017_ICCV} on AUC.

\begin{figure}[t]
  \centering
    \includegraphics[width=0.95\textwidth]{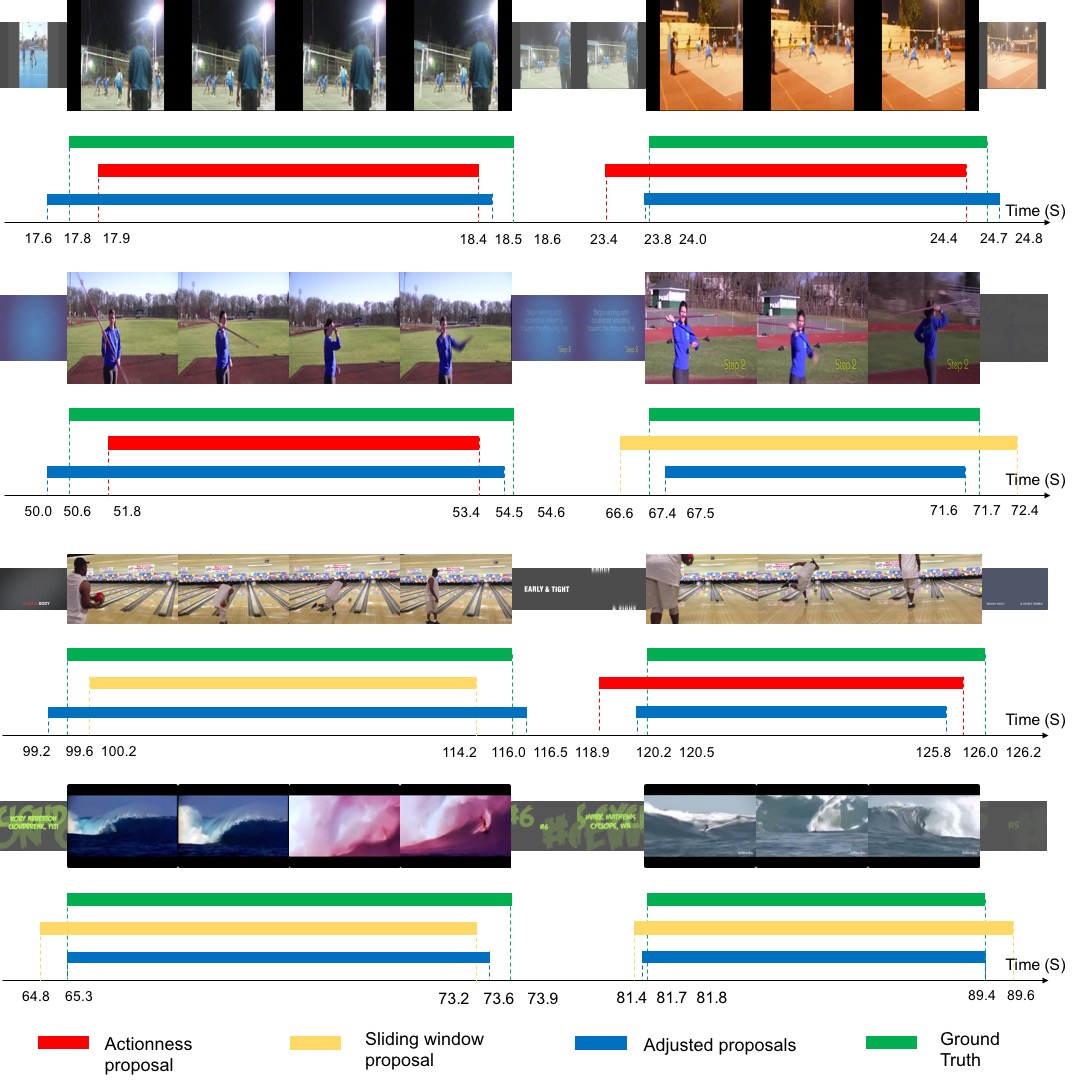}
    \caption{Visualization of temporal action proposals generated by CTAP. First two rows represent 4 temporal action proposals from 2 videos in THUMOS-14. Last two rows represent 4 temporal action proposals from 2 videos in ActivityNet v1.3.}
      \label{fig: demo}
\end{figure}

\textbf{Evaluation of PATE.} Based on TAR, we further explore the function of PATE complementary filtering. 
We evaluate three different models: (1) sliding window proposals with TAR (SW-TAR) (2) actioness proposals with TAR (TAG-TAR) (3) PATE Complementary proposals with TAR (our full model, CTAP). 
Different models' performances of AR@100 and AUC are reported in Table~\ref{tbl: anet tar eval}. 
CTAP achieves consistently better performance of AR@100 and AUC compared to SW-TAR and TAG-TAR, which shows its advantage of selecting complementary proposals from sliding windows to fill the omitted ones in actionness proposals.

\textbf{Comparison with state-of-the-art methods.} CTAP is compared with state-of-the-art methods on ActivityNet v1.3 validation set by the Average Recall at top 100 ranked proposals (AR@100) and area under AR-AN curve (AUC). 
In Table~\ref{tbl: anet tar eval}, we find CTAP achieves 2.60\% and 1.32\% increase in AR@100 compared with state-of-the-art methods MSRA~\cite{yao2017msr} and  Prop-SSAD~\cite{lin2017temporal} respectively. 


\begin{table}[t]
\centering
\caption{Generalization evaluation of CTAP on Activity Net v1.3 (validation set) in terms of AR@100 and AR-AN under curve area.}\label{tbl: anet generalization}
\begin{tabular}{l*2c}
\toprule
{} &\ Seen (100 classes)\ &\ Unseen (100 classes)\ \\\midrule
AR@100\ &\ 74.06\ &\ 72.51\ \\ 
AR-AN\ &\ 66.01\ &\ 64.92\ \\\bottomrule
\end{tabular}
\end{table}

\textbf{Generalization ability of proposals.} 
We evaluate the generalization ability of CTAP on ActivityNet v1.3 validation set. Following the setting of~\cite{xiong2017pursuit}, we evaluate the AR@100 and AR-AN under curve area (AUC) for 100 seen classes and unseen classes respectively. 
In Table~\ref{tbl: anet generalization}, we observe that CTAP achieves better performance on 100 seen classes. On unseen 100 classes, there is only a slight drop in AR@100 and AUC, which shows the generalizability of CTAP.

\subsection{Qualitative Results}
We further visualize some temporal action proposals generated by CTAP. As shown in Fig.~\ref{fig: demo}, CTAP is able to select most suitable initial proposals from actionness proposals or sliding windows, and then adjust their temporal boundaries more precisely.

\section{Conclusion}
Previous methods for temporal action proposal generation can be divided to two groups: sliding window ranking and actionness score grouping, which are complementary to each other: sliding windows uniformly cover all segments in videos, but the temporal boundaries are imprecise; actionness score based method may have more precise boundaries but it may omit some proposals when the quality of actioness scores is low. We propose a novel Complementary Temporal Action Proposal (CTAP) generator, which could collect high quality complementary proposals from sliding windows and actionness proposals. A temporal convolutional network for proposal ranking and boundary adjustment is also designed. CTAP outperforms state-of-the-art methods by a large margin on both THUMOS-14 and ActivityNet 1.3 datasets. Further experiments on action detection show consistent large performance improvements.  

\section*{Acknowledgements}
This research was supported, in part, by the Office of Naval Research under grant N00014-18-1-2050 and by an Amazon Research Award. 

%
%
%
\bibliographystyle{splncs04}
\bibliography{egbib}
%




\end{document}